\documentclass[acmtog,nonacm]{acmart}
\makeatletter
\let\@authorsaddresses\@empty
\makeatother

\AtBeginDocument{%
  }

\usepackage{booktabs}
\usepackage{multirow}
\usepackage{makecell}
\definecolor{magenta}{rgb}{1,0,1} 

\def\ie{\textit{i.e.}}
\def\eg{\textit{e.g.}}

\begin{document}

\title{Voyager: Long-Range and World-Consistent Video Diffusion for Explorable 3D Scene Generation}

\author{Tianyu Huang}
\authornote{Both authors contributed equally to this research.}
\affiliation{%
  \institution{Harbin Institute of Technology}
  \country{China}
}

\author{Wangguandong Zheng}
\authornotemark[1]
\affiliation{%
  \institution{Southeast University}
  \country{China}}

\author{Tengfei Wang}
\authornote{\footnotemark[1]Corresponding author.}  
\affiliation{%
  \institution{Tencent Hunyuan}
  \country{China}}

\author{Yuhao Liu}
\affiliation{%
  \institution{City University of Hong Kong}
  \country{China}}

\author{Zhenwei Wang}
\affiliation{%
  \institution{City University of Hong Kong}
  \country{China}}

\author{Junta Wu}
\affiliation{%
  \institution{Tencent Hunyuan}
  \country{China}}

\author{Jie Jiang}
\affiliation{%
  \institution{Tencent Hunyuan}
  \country{China}}

\author{Hui Li}
\affiliation{%
  \institution{Harbin Institute of Technology}
  \country{China}}

\author{Rynson W.H. Lau}
\affiliation{%
  \institution{City University of Hong Kong}
  \country{China}}

\author{Wangmeng Zuo}
\authornotemark[2]
\affiliation{%
  \institution{Harbin Institute of Technology}
  \country{China}}

\author{Chunchao Guo}
\affiliation{%
  \institution{Tencent Hunyuan}
  \country{China}}

\renewcommand{\shortauthors}{Tianyu Huang, Wangguandong Zheng, Tengfei Wang, Yuhao Liu et al.}

\begin{teaserfigure}
  \centering
  \vspace{-0.5em}
  \includegraphics[width=0.92\textwidth]
  {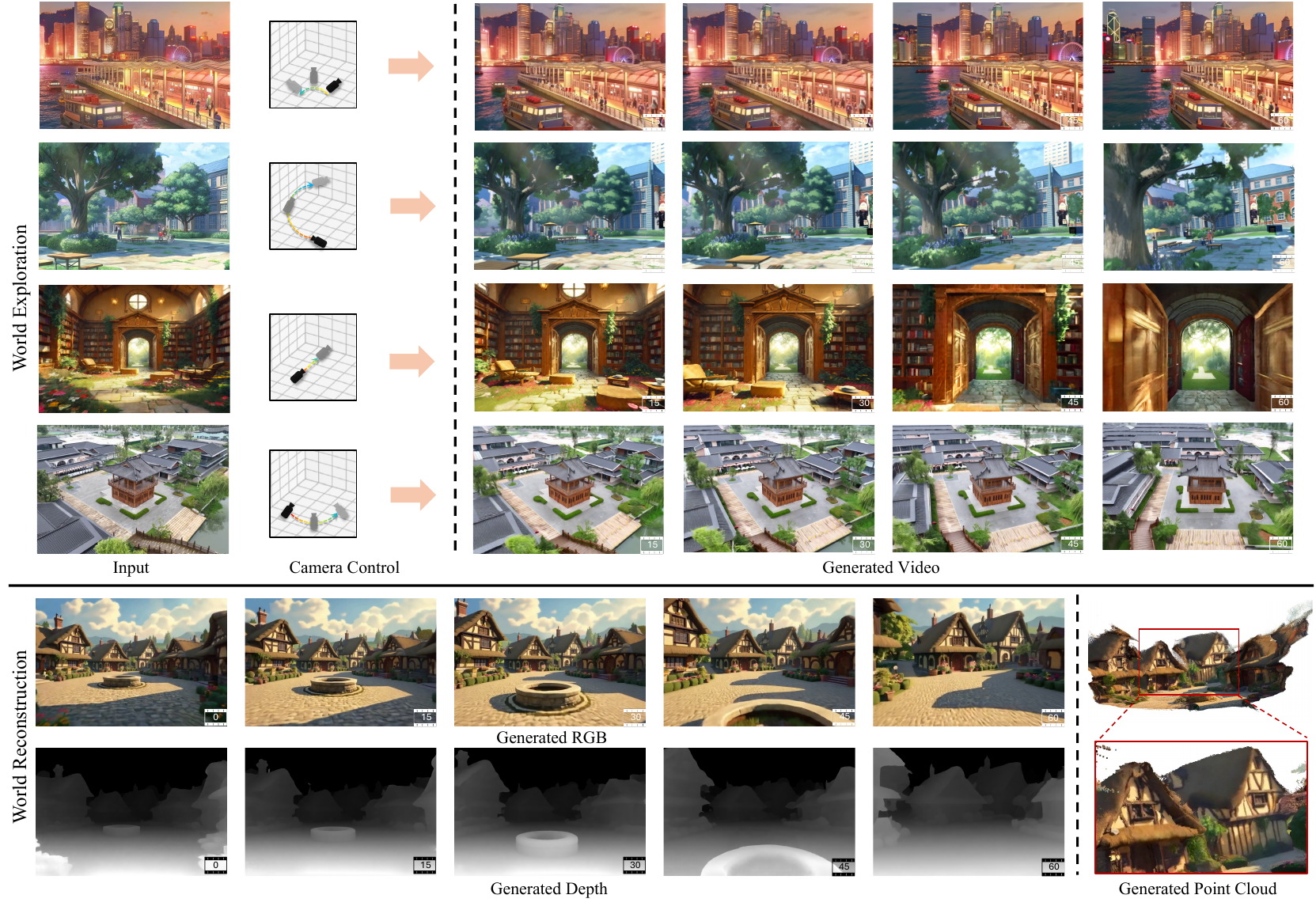}
  \vspace{-1.5em}
  \caption{\emph{Voyager} is a world-consistent video generation and reconstruction framework. Up: Voyager can generate 3D-consistent scene videos for world exploration following custom camera trajectories. Bottom: Voyager jointly generates aligned depth and RGB video for effective and direct 3D reconstruction.}
  \Description{.}
  \vspace{-0.5em}
  \label{fig:teaser}
\end{teaserfigure}

\begin{abstract}
Real-world applications like video gaming and virtual reality often demand the ability to model 3D scenes that users can explore along custom camera trajectories. While significant progress has been made in generating 3D objects from text or images, creating long-range, 3D-consistent, explorable 3D scenes remains a complex and challenging problem. In this work, we present \textbf{\emph{Voyager}}, a novel video diffusion framework that generates world-consistent 3D point-cloud sequences from a single image with user-defined  camera path. 
Unlike existing approaches, Voyager achieves end-to-end scene generation and reconstruction with inherent consistency across frames, eliminating the need for 3D reconstruction pipelines (\textit{e.g.,} structure-from-motion or multi-view stereo). 
Our method integrates three key components:  1) \textbf{World-Consistent Video Diffusion}: A unified architecture that jointly generates aligned RGB and depth video sequences, conditioned on existing world observation to ensure global coherence 2) \textbf{Long-Range World Exploration}: An efficient world cache with point culling and an auto-regressive inference with smooth video sampling for iterative scene extension with context-aware consistency, and 3) \textbf{Scalable Data Engine}: A video reconstruction pipeline that automates camera pose estimation and metric depth prediction for arbitrary videos, enabling large-scale, diverse training data curation without manual 3D annotations. Collectively, these designs result in a clear improvement over existing methods in visual quality  and geometric accuracy, with versatile applications. See more at \href{https://voyager-world.github.io}{\textcolor{magenta}{https://voyager-world.github.io}}.
\end{abstract}

\maketitle

\section{Introduction}
The creation of high-fidelity, explorable 3D scenes that users can navigate seamlessly, powers broad applications ranging from video gaming and film production to robotic simulation.  Yet, traditional workflows for constructing such 3D worlds remain bottlenecked by manual effort, requiring painstaking layout design, asset curation, and scene composition. While recent data-driven methods~\cite{xiang2024structured,zhao2025hunyuan3d,meng2024lt3sd,xie2024citydreamer,liu2024citygaussian} have shown promise in generating objects or simple scenes, their ability to scale to complex scenes is limited by the scarcity of high-quality 3D scene data.
 This gap highlights the need for frameworks that enable scalable generation of user-navigable virtual worlds with 3D consistency.

\begin{figure}
    \centering
    \includegraphics[width=0.98\linewidth]{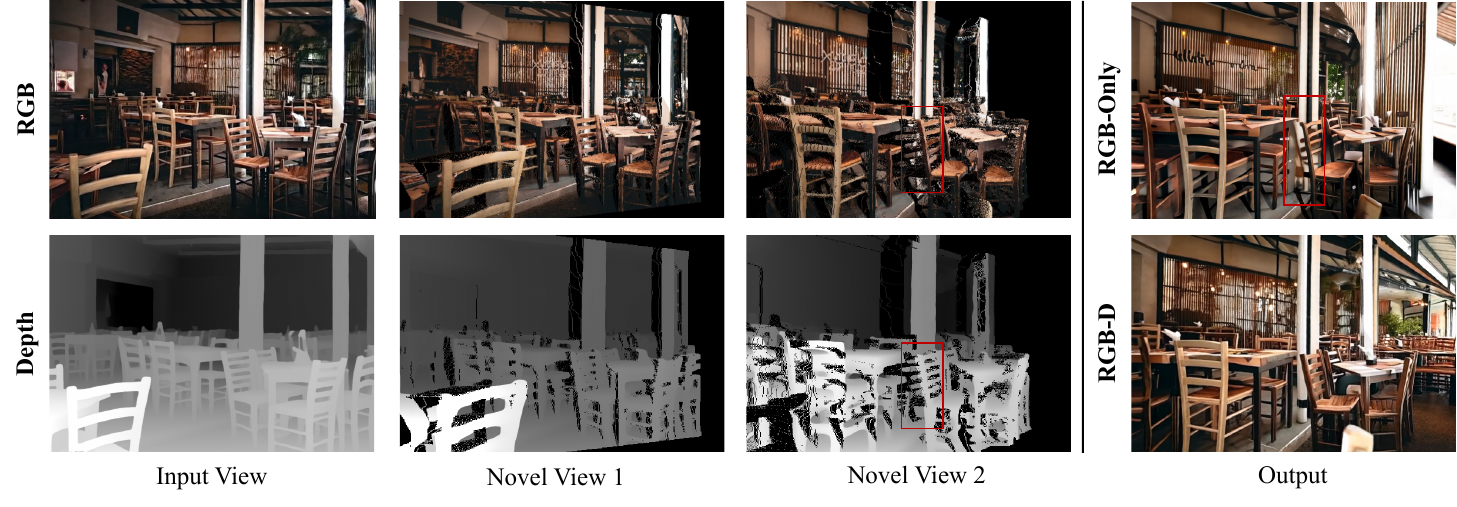}
    \vspace{-1.em}
    \caption{Partial RGB images and partial depth maps rendered from point clouds at different frames. In scenarios involving complex occlusion relationships, partial RGB images often exhibit significant visual artifacts. In contrast, partial depth maps can accurately represent occlusions.}
    \label{fig:motivation}
    \vspace{-1em}
\end{figure}

Recently, a growing number of works~\cite{gao2024cat3d,he2024cameractrl,wang2024motionctrl,yu2024viewcrafter,Ma2025See3D,zhou2025stable,ren2025gen3c,chen2025flexworld} have explored the use of novel view synthesis (NVS) and video generation as alternative paradigms for world modeling. These methods, while demonstrating impressive capabilities in generating visually appealing and semantically rich content, still face several challenges. \textbf{1) Long-Range Spatial Inconsistency.} Due to the absence of explicit 3D structural grounding, they often struggle to maintain spatial consistency and coherent viewpoint transitions during the generation process, especially when generating videos with long-range camera trajectories.
\textbf{2) Visual Hallucination}. While several works~\cite{ren2025gen3c,chen2025flexworld} have attempted to leverage 3D conditions to enhance geometric consistency, they typically rely on partial RGB images as guidance, \ie, novel-view images rendered from point clouds reconstructed with input views. However, such representation may introduce significant visual hallucinations in complex scenes, such as the incorrect occlusions in Figure.~\ref{fig:motivation}, which may introduce inaccurate supervision during training.
\textbf{3) Post-hoc 3D Reconstruction.} While these approaches can synthesize visually satisfying content, post-hoc 3D reconstructions are still required to obtain usable 3D content. This process is time-consuming and inevitably introduces geometric artifacts~\cite{weber2024toon3d}, making it inadequate for real-world applications.

To address these challenges, we introduce \emph{Voyager}, a framework designed to synthesize long-range, world-consistent RGB-D(epth) videos from a single image and user-specified camera trajectories. At the core of Voyager is a novel \textbf{world-consistent video diffusion}  that utilizes an expandable world caching mechanism to ensure spatial consistency and avoids visual hallucination. Starting from an image, we construct an initial world cache by unprojecting it into 3D space with a depth map. This 3D cache is then projected into target camera views to obtain partial RGB-D observations, which guides the diffusion model to maintain coherence with the accumulated world state. Crucially, the generated frames are fed back to update and expand the world cache, creating a closed-loop system that supports arbitrary camera trajectories while maintaining geometric coherence.

Unlike methods~\cite{yu2024viewcrafter,Ma2025See3D,ren2025gen3c,chen2025flexworld} relying only on RGB conditioning, Voyager explicitly leverages depth information as a spatial prior, enabling more accurate 3D consistency during video generation. By simultaneously generating aligned RGB and depth sequences, our framework supports direct 3D scene reconstruction without requiring additional 3D reconstruction steps like structure-from-motion.

Despite promising performance, diffusion models struggle to generate long videos in a single pass. To enable \textbf{long-range world exploration}, we propose world caching scheme and smooth video sampling for auto-regressive scene extension. Our world cache accumulates and maintains point clouds from all previously generated frames, expanding as video sequences grow. To optimize computational efficiency, we design a point culling method to detect and remove redundant points with real-time rendering, minimizing memory overhead. Leveraging cached point clouds as a proxy, we develop a smooth sampling strategy that auto-regressively extends video length while ensuring smooth transitions between clips.

Training such a model requires large-scale videos with accurate camera poses and depth, but existing datasets often lack these annotations. To address this, we introduce a data engine for \textbf{scalable video reconstruction} that automatically estimates camera poses and metric depth for arbitrary scene videos. With metric depth estimation, our data engine ensures consistent depth scales across diverse sources, enabling high-quality training data generation. Using this pipeline, we compile a dataset of over 100,000 video clips, combining real-world captures and synthetic Unreal Engine renders.

Extensive experiments demonstrate the effectiveness of Voyager in scene video generation and 3D world reconstruction.  Benefiting from joint depth modeling, our results in Figure~\ref{fig:teaser} exhibit more coherent geometry, which not only enable direct 3D reconstruction but also support infinite world expansion while preserving the original spatial layout. Additionally, we explore applications such as 3D generation, video transfer, and depth estimation, further showcasing the potential of Voyager in advancing spatial intelligence. 

Our contributions can be summarized as:
\begin{itemize}
    \item We introduce Voyager, a world-consistent video diffusion model for scene generation. To the best of our knowledge, Voyager is the first video model that jointly generates RGB and depth sequences with given camera trajectories.
    \item We propose an efficient world caching scheme and auto-regressive video sampling approach, extending Voyager to world reconstruction and infinite world exploration.
    \item We propose a scalable video data engine for camera and metric depth estimation, with over 100,000 training pairs prepared for the video diffusion model.
\end{itemize}
\section{Related Work}
\subsection{Camera-Controllable View Generation}
Existing camera-controllable generation models can be categorized into three types: novel view synthesis~\cite{kerbl20233d,mildenhall2021nerf,wu2024reconfusion,hong2023lrm} produces new viewpoints through multi-view reconstruction. These methods rely on dense viewpoints and struggle to handle single-view inputs. The second method~\cite{guo2023animatediff,liu2023zero1to3,wang2024motionctrl,he2024cameractrl,zhou2025stable} implicitly incorporates camera parameters into the model, training it to generate images from the corresponding viewpoints, but often suffers from viewpoint inconsistency. The third method~\cite{seo2024genwarp,Ma2025See3D,ren2025gen3c,chen2025flexworld} leverages point clouds obtained by warping the input view as conditions for novel view generation, significantly improving spatial consistency. However, the warped images still contain artifacts that negatively affect model training. In this work, we introduce warping depth as an additional conditioning input and generate both RGB and depth content.

\subsection{Long-Range Video Generation}
Current video models are limited in their ability to generate long videos in a single pass. To extend video length, existing research explores training-free methods~\cite{wang2023gen,lu2024freelong}, hierarchical strategies~\cite{yin2023nuwa,he2022latent}, and auto-regressive frameworks~\cite{yin2024slow,henschel2024streamingt2v}. However, the first two approaches cannot scale to infinitely long videos, while the auto-regressive strategy relies on memory caches that struggle to retain information from distant past frames. To address this limitation, we propose world cache with point culling in this work that efficiently preserves spatial information and enables the generation of arbitrarily long videos with smooth video sampling in an auto-regressive inference.

\begin{figure*}
    \centering
    \includegraphics[width=0.9\textwidth]{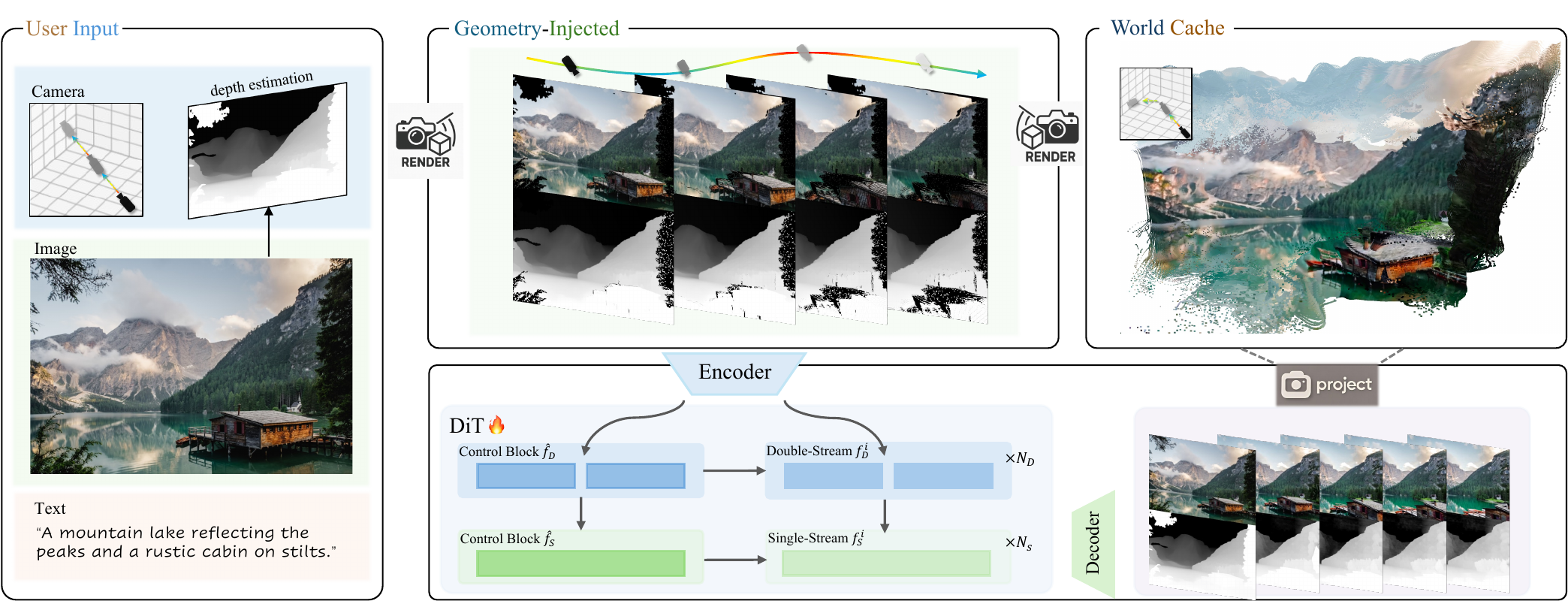}
    \vspace{-1em}
    \caption{Overview of Voyager: Given the input image and camera trajectories, we first render partial RGB images and depth maps for each viewpoint as the condition for video generation. Our world-consistent video diffusion model is trained to generate RGB-D frames simultaneously, thus supporting the direct reconstruction of the 3D world. The projected points are store in our world cache efficiently, which can be rendered as condition for the next round generation.}
    \vspace{-1.5em}
    \label{fig:method}
\end{figure*}

\section{Preliminaries of Video Diffusion Models}
Diffusion models learn to denoise a data distribution $p(x)$ through an iterative process, in which a forward diffusion process gradually adds noise to the data $\mathbf{x}_0 \sim p(\mathbf{x}_0)$, and a reverse process learns to recover $\mathbf{x}_0$ from the noisy data $\mathbf{x}_t$.

In the context of video generation, diffusion models are extended to learn temporal dynamics by incorporating 3D convolutional architectures~\cite{tran2015learning,yu2023language} and attention mechanisms~\cite{brooks2024video,zhang2023dinet}. 
To reduce the computation cost, latent diffusion~\cite{rombach2022high,blattmann2023align} is widely used to compress the video to a low-dimensional latent space.

In this work, our video model is based on Hunyuan-Video~\cite{kong2024hunyuanvideo}. Formally, given an input text prompt $y$ and a ground-truth video sequence $[I_0,..., I_{T-1}] \in \mathbb{R}^{T \times 3 \times H \times W}$, the model first extracts the video latent $\mathbf{z}_0$ with shape $(\frac{T}{c_t}+1) \times C \times \frac{H}{c_s} \times \frac{W}{c_s}$ by a 3D-VAE, where $c_t$ and $c_s$ denote the compression rate for the temporal and spatial axis.
To train a denoising model $\theta$, noisy latent $\mathbf{z}_t$ is then fed to a full-attention DiT~\cite{Peebles2022DiT,li2024hunyuan} model, which follows the strategy of "Dual-stream to Single-stream" hybrid model~\cite{flux2024}. Patched video and text latents are processed independently in dual-stream Transformer blocks $f_D^i$, while in the second phase, these latents are concatenated in single-stream blocks $f_S^i$. To further support image-conditioned video generation, the latent feature of the input image is concatenated to $\mathbf{z}_t$ channel-wise. The training objective is to predict the velocity $\mathbf{u}_t = d\mathbf{z}_t/dt$ by minimizing the mean squared error between the estimated velocity $\hat{\mathbf{u}}_t$ and the ground-truth $\mathbf{u}_t$. Finally, the latent $\mathbf{z}_0$ is recovered by the first-order Euler ordinary differential equation (ODE) solver, and the video $v$ is reconstructed by the 3D-VAE decoder.

\section{Methodology: Voyager}
Given an image $I_0 \in \mathbb{R}^{3 \times H \times W}$, our goal is to create an explorable world based on a user-defined camera trajectory. However, there is a gap between video generation and 3D world modeling, which mainly stems from three aspects: (1) the inconsistency of long-range video extension, (2) the hallucination of visual conditions for video generation, and (3) the incapability of reconstructing the world from video outputs. To address these issues, we propose \textbf{\emph{Voyager}}, a world-consistent video generation framework that can directly produce rgb-depth frames with corresponding camera parameters for long-range world exploration. In this section, we first introduce a geometry-injected frame condition to compensate for perceptual hallucination under visual conditions. (Sec.~\ref{subsec:cond}). With this input condition, we propose a depth-fused video diffusion model to ensure spatial consistency and context-based blocks to enhance its viewpoint control (Sec.~\ref{subsec:rgbd}). For 3D world reconstruction and long-range exploration, we propose world caching with point culling and smooth video sampling in the auto-regressive inference (Sec.~\ref{subsec:infer}). We further propose a scalable video data engine to prepare camera and metric depth for the training of the above model (Sec.~\ref{subsec:train}).

\subsection{Geometry-Injected Frame Condition}\label{subsec:cond}
For the control of video viewpoint, camera parameter~\cite{zhou2025stable,bai2025recammaster} is a straightforward condition, but this implicit condition is nontrivial to the training of video models.
Recent works~\cite{Ma2025See3D,ren2025gen3c,chen2025flexworld} attempt to reconstruct the point cloud $p \in \mathbb{R}^{N \times 6}$ from videos as an explicit control, where $N$ is the number of points and each point is represented by 6D coordinates $(x,y,z,r,g,b)$. The warped RGB condition $\hat{I}_v$ for a novel view $v$ can then be rendered according to the camera, which is a partial image with blank regions.

Nonetheless, such a partial RGB image is insufficient to ensure spatial consistency, \eg, complex occlusion relationships in a scene may lead to visual hallucinations. To enforce spatially consistent control during training, we introduce an additional geometric condition partial depth map, which is aligned with the partial RGB image. Specifically, we first estimate the depth map $D_k$ and corresponding camera parameters $c_k$ for each frame $I_k$ of the video. Since only the first frame is visible in video inference, we create a point cloud $p_0$ by projecting $D_0$ with $c_0$. For the $k$-th frame, its partial image $\hat{I}_k$ and partial depth $\hat{D}_k$ are acquired by masking the invisible region with the rendering mask $M_k=\mathrm{render}(p_0, c_k)$.

\subsection{World-Consistent Video Diffusion}\label{subsec:rgbd}
Conditioned with partial RGB and depth maps, our intention is to generate plausible content for the invisible regions, ensuring consistency with the spatial information provided by the partial conditions.

For this purpose, the common practice~\cite{flux2024,ren2025gen3c} is to concatenate the condition latents $\mathbf{z}_\mathrm{rgb}$ and $\mathbf{z}_\mathrm{depth}$ with original noisy latents $\mathbf{z}_t$ along the \textbf{channel} axis and project the concatenated latents back to the Transformer dimension via the patch-embedding layer $f_{\mathrm{emb}}$: $\mathbf{z}'_{t,0} = f_{\mathrm{emb}}(\mathrm{concat}(\mathbf{z}_t, \mathbf{z}_\mathrm{rgb}, \mathbf{z}_\mathrm{depth}))$. 
Then, the projected latents $\mathbf{z}'_{t,0}$ are fed to double-stream and single-stream blocks sequentially, which is formulated as,

\begin{equation}\label{eq:double}
    \mathbf{z}'_{t,i}, \mathbf{z}'_{y,i} = f_D^i(\mathbf{z}'_{t,i-1}, \mathbf{z}'_{y,i-1}, t), i=1,...,N_D,
\end{equation}

\begin{equation}\label{eq:single}
    \mathbf{z}''_{t,i} = f_S^i(\mathbf{z}''_{t,i-1}, t), i=1,...,N_S,
\end{equation}
where $\mathbf{z}'_y$ is the text latents. $N_D$ and $N_S$ denote the block number of each stream. $\mathbf{z}''_{t,0}$ is initialized as the concatenation of $\mathbf{z}'_{t,N}$ and $\mathbf{z}'_{y,N}$.

Although the video model can best preserve the pre-trained parameters in this way, the spatial conditions is only used in channel-wise. The missing parts in our partial maps can range from small cracks to large blank areas, depending on the extent of the viewpoint change. This trivial solution struggles to handle variable situations.

\noindent\textbf{Depth-Fused Video Generation.}
Instead of relying solely on partial depth as the input condition for completing the missing regions in the RGB frames, we propose to simultaneously generate both complete RGB and depth frames.
As a result, the video model can take advantage of DiT's full-attention structure, allowing for the interaction of visual and geometric information at the pixel level. 
To this end, we concatenate the rgb and depth images along the \textbf{height} axis as $\mathcal{I}_k=[I_k, \Phi, D_k]_h$, as well as condition maps $\hat{\mathcal{I}}_k=[\hat{I}_k, \Phi, \hat{D}_k]_h$ and masks $M_k=[M_k, \Phi, M_k]_h$. Here, we add a placeholder row $\Phi$ between the rgb and depth images to help the model separate these two types of content. The new video latents are presented as $\mathbf{z}'_{t,0} = f_{\mathrm{emb}}(\mathrm{concat}(\mathbf{z}_t, \hat{\mathbf{z}}_0, m))$, where $\hat{\mathbf{z}}_0$ is the latent of $[\hat{\mathcal{I}}_k]_{k=0}^{T-1}$ and $m$ is the down-sampled map of $[M_k]_{k=0}^{T-1}$ via max-pooling. Accordingly, $\mathbf{z}'_{t,0}$ is fed to the diffusion model similar to Eq.~\ref{eq:double}-\ref{eq:single}. The diffusion model is thus trained to generate rgb-depth video frames.

\noindent\textbf{Context-Based Control Enhancement.} 
The above concatenation mechanism incorporates conditional information only at the input of the DiT model, leading to weak enforcement of the geometric conditions and resulting in misalignment between generated frames and input conditions. 

To enhance the geometric-following capabilities, following~\cite{bian2025videopainter}, we further inject the diffusion model with lightweight modules. Concretely, we replicate the first block from the double-stream and single-stream modules as the Control blocks $\hat{f}_D$ and $\hat{f}_S$. Given the input video latent $\mathbf{z}'_{t,0}$, we have the following operations for each Transformer block $i$:
\begin{equation}
    \mathbf{z}_D = \hat{f}_D(\mathbf{z}'_{t,0}), \mathbf{z}_S = \hat{f}_S(\mathbf{z}_D),
\end{equation}
\begin{equation}
    \mathbf{z}'_{t,i} = \mathbf{z}'_{t,i} + l_D(\mathbf{z}_D), \mathbf{z}''_{t,i} = \mathbf{z}''_{t,i} + l_S(\mathbf{z}_S),
\end{equation}
where $l_D$ and $l_S$ are zero-initialized linear layers. Early-stage latent features preserve more contextual information, so that the integration into each block can strengthen pixel-level controllability.

\subsection{Long-Range World Exploration}\label{subsec:infer}
For long-range or even infinite video generation, auto-regressive is a natural choice. This paradigm recursively generates future frames or clips based on previously generated content, maintaining temporal continuity over time. However, due to the limited memory capacity of video diffusion models, auto-regressive methods are often restricted to conditioning on only a few preceding frames or clips. This limited context leads to inevitable information loss, making it fundamentally infeasible to retain and propagate the full scene history. In contrast to previous auto-regressive methods, Voyager exploits point-cloud conditions for generation, which is a scalable representation to store the whole history information. To enable infinite generation, we propose world caching with point culling to efficiently store spatial information and adopt smooth video sampling to ensure the consistency of consecutive clips.

\begin{table}[t]
\centering
\caption{Quantitative comparison of novel view synthesis on \textit{RealEstate10K}.}
\label{tab:video}
\vspace{-1em}
\setlength{\tabcolsep}{12pt}
\scalebox{0.9}{
\begin{tabular}{lccc}
\toprule
\multirow{1}{*}{\textbf{Method}} & \textbf{PSNR} $\uparrow$ & \textbf{SSIM} $\uparrow$ & \textbf{LPIPS} $\downarrow$\\
\midrule
SEVA & 16.648&  0.613& 0.349 \\
\hline
ViewCrafter &  16.512&  0.636& 0.332 \\
See3D & 18.189&  \underline{0.694}& 0.290 \\
FlexWorld & \underline{18.278}&  0.693& \underline{0.281} \\
\hline
\cellcolor{gray!15}\textbf{Voyager} & \cellcolor{gray!15}\textbf{18.751} & \cellcolor{gray!15}\textbf{0.715} & \cellcolor{gray!15}\textbf{0.277} \\
\bottomrule
\end{tabular}
}
% \vspace{-2em}
\end{table}

\noindent\textbf{World Caching with Point Culling.} With input camera parameters and corresponding RGB-D video frames, point clouds can be projected to 3D space as $\hat{p}\in\mathbb{R}^{(T\times H \times W) \times 3}$, where $T$ is the number of total frames. As the video continues to extend, the number of points can easily grow to millions, posing significant challenges in terms of memory and computational efficiency. To address that, we propose to maintain a world cache, which eliminates redundant points while preserving essential geometric information. Specifically, we incrementally add new points to the cache on a per-frame basis: given the accumulated point cloud $\hat{p}$ from previous frames, we render a visibility mask $M = \text{render}(\hat{p}, c_i)$ from the current camera view $c_i$. Points in the invisible regions are added to $\hat{p}$ first. For the visible regions, if the angle between the surface normal of existing points and the current view direction exceeds 90 degrees, the new point is also updated into the cache, because these existing points cannot be seen at the current viewpoint. 
This strategy reduces the number of stored points by approximately 40\% and avoids noise accumulation caused by multi-frame aggregation.

\noindent\textbf{Smooth Video Sampling.} Conditioned on the above world cache, our video model can access the complete spatial information from previous frames. However, although each independently generated video clip is spatially consistent, there can still be color discrepancies, making them unsuitable for direct concatenation. 
We adopt two strategies to ensure smoother transitions between adjacent clips. (1) We first divide the input video into overlapping segments, where the length of the overlapping region is half of one segment. For each segment, the overlapping region is initialized with the generated results from the previous segment, serving as the noise initialization for the current segment's overlap region. 
(2) After completing inference for the consecutive two segments, we apply averaging across the overlapping regions and introduce a light-level noise injection to the merged segments. A final round of denoising is then performed to refine transitions.
In this way, we ensure the efficient generation of multiple clips while maintaining visual consistency across consecutive video frames. 

\subsection{Scalable Video Data Engine}\label{subsec:train}
Training such a video model demands large-scale video frames with corresponding camera parameters and depth maps. We carefully curate over 100,000 video clips from both real-captured videos and 3D renderings, and propose a scalable video data engine to automatically annotate required 3D information for arbitrary scene videos.

\noindent\textbf{Data Curation.}
We selected two open-source real-world datasets, \ie, RealEstate~\cite{zhou2018stereo} and DL3DV~\cite{ling2024dl3dv} for the training. RealEstate contains 74,766 video clips related to real estate scenes, primarily featuring indoor home scenes, along with some outdoor environments. DL3DV provides 10K real-scene videos, but most of them suffer from rapid or shaky camera movements. We curate 3,000 high-quality videos from this dataset and segment them into approximately 18,000 video clips. Additionally, to increase the diversity of generation content, we collected 1,500 Unreal Engine scene models and rendered over 10,000 video samples to augment the dataset. In the end, we collected over 100,000 video clips from these datasets.

\noindent\textbf{Data Annotation.}
Accurate camera parameters and depth are crucial for model training, but RealEstate and DL3DV do not provide such ground-truth data. Existing methods~\cite{chen2025flexworld,ren2025gen3c,schwarz2025recipe} adopt dense stereo models~\cite{teed2021droid} to prepare training pairs, 
struggling to produce geometrically consistent depth.
We propose a more robust data processing engine. Specifically, we first use VGGT~\cite{wang2025vggt} to estimate camera parameters and depth for all video frames. The depth estimated by VGGT is not accurate enough, but it is aligned with camera poses. To further improve the depth estimation, we then employ MoGE~\cite{wang2024moge} as a robust depth estimator and align the two depth maps with least squares optimization. 

Finally, since our UE data provides metric depth values, we need to align all the estimated depth to a standard scale. We estimate the metric depth range of the scene using Metric3D~\cite{hu2024metric3d} and map the previous depths into this range. This way, we can automatically annotate camera and depth for videos from any source.

\begin{table}[t]
\centering
\caption{Quantitative comparison of Gaussian Splattig reconstruction on \textit{RealEstate10K}. Baselines require additional reconstruction step~\cite{wang2025vggt}, while Voyager  performs better with our generated depth.}
\label{tab:quantitative_comparison}
\vspace{-1em}
\scalebox{0.9}{
\begin{tabular}{lcccc}
\toprule
\multirow{1}{*}{\textbf{Method}} & \textbf{Post  Rec.} & \textbf{PSNR} $\uparrow$ & \textbf{SSIM} $\uparrow$ & \textbf{LPIPS} $\downarrow$ \\
% & \textbf{Latency} $\downarrow$ \\
\midrule
SEVA & VGGT & 15.581&  0.602& 0.452\\
\hline
ViewCrafter & VGGT & 16.161&  0.628& 0.440 \\
See3D & VGGT & 16.764&  0.633& 0.440\\
FlexWorld & VGGT & 17.623&  0.659& 0.425 \\
\hline
\cellcolor{gray!15}\textbf{Voyager} & \cellcolor{gray!15}VGGT & \cellcolor{gray!15}\underline{17.742} & \cellcolor{gray!15}\underline{0.712} & \cellcolor{gray!15}\underline{0.404} \\
\cellcolor{gray!15}\textbf{Voyager} & \cellcolor{gray!15}- & \cellcolor{gray!15}\textbf{18.035} & \cellcolor{gray!15}\textbf{0.714} & \cellcolor{gray!15}\textbf{0.381}\\
\bottomrule
\label{tab:scene}
\end{tabular}}
\vspace{-2.5em}
\end{table}
\begin{figure*}
    \centering
    \includegraphics[width=0.8\textwidth]{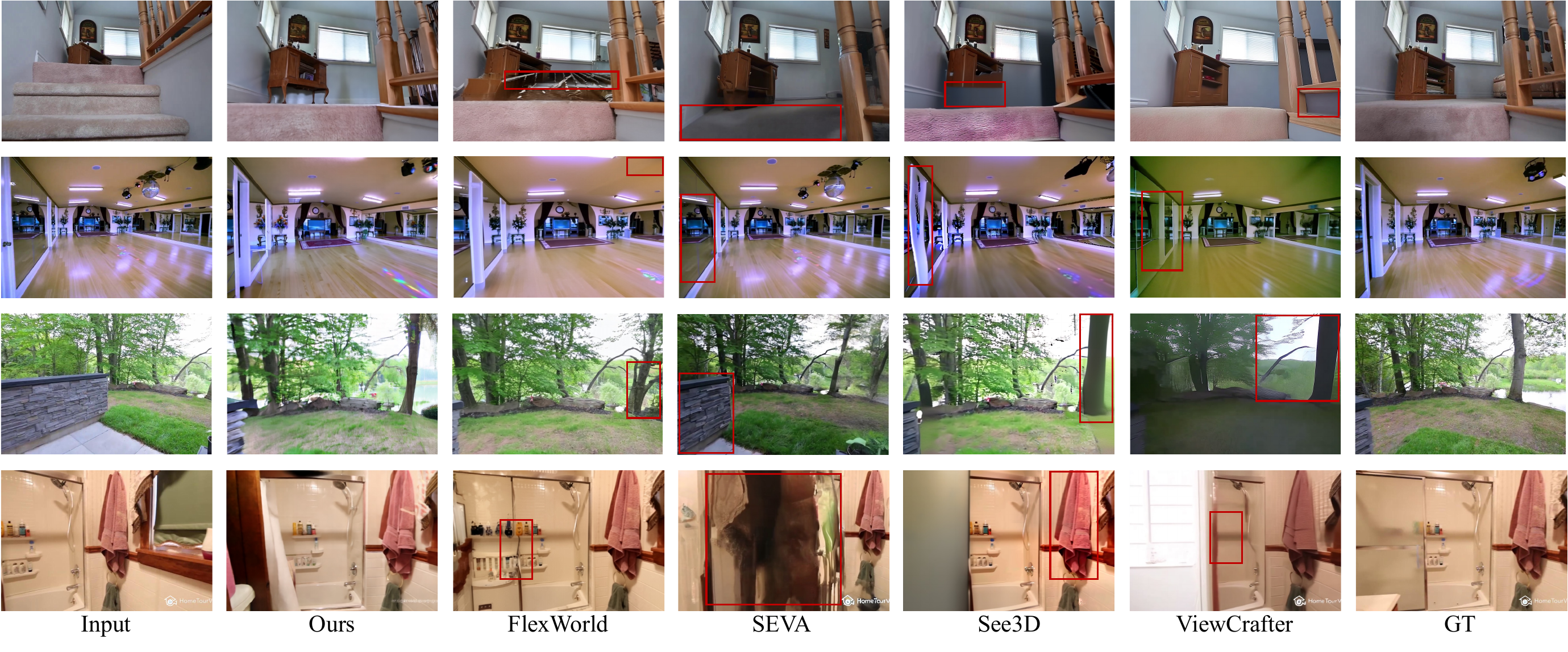}
    \vspace{-1.em}
    \caption{Qualitative results on video generation. Compared to the baselines, our model can generate a more reasonable unseen region and meanwhile preserve the content in the input view.}
    \label{fig:video}
    \vspace{-1.5em}
\end{figure*}

\section{Experiments}
\subsection{Video Generation}\label{subsec:video}
We evaluate the video generation quality of Voyager by comparing four open-source camera-controllable video generation methods on image-to-video generation, including SEVA~\cite{zhou2025stable}, ViewCrafter~\cite{yu2024viewcrafter}, See3D~\cite{Ma2025See3D}, and FlexWorld~\cite{chen2025flexworld}. Among these methods, ViewCrafter, See3D, and FlexWorld control the viewpoints with point cloud conditions, which are similar to our method. SEVA directly takes camera parameters as input conditions.

\noindent\textbf{Dataset and Metrics.} We randomly select 150 video clips from the test set of RealEstate~\cite{zhou2018stereo} as our test dataset. Since the video clips do not provide ground-truth cameras, we estimate the camera parameters and depth maps with the same pipeline in our data engine. To evaluate the visual quality of generated videos, we adopt PSNR, SSIM, and LPIPS to measure the similarity between the generated frames and the ground truth.

\noindent\textbf{Results.} We report the quantitative results on Table~\ref{tab:video}. Our method outperforms all the baselines, demonstrating the high generation quality of our video model. The qualitative comparison in Figure~\ref{fig:video} also showcases our capability of generating photorealistic videos. Especially in the last case of Figure~\ref{fig:video}, only our method can preserve the details of products in the input image. However, other methods are prone to generating artifacts, \eg, in the first example of Figure~\ref{fig:video}, these methods fail to provide reasonable predictions when the camera movement is too large.

\vspace{-3mm}
\subsection{Scene Generation}\label{subsec:scene}
To evaluate the quality of scene generation, we further compare the quality of scene reconstruction with generated videos based on Sec.~\ref{subsec:video}. Since the compared baselines only produce RGB frames, we first exploit VGGT~\cite{wang2025vggt} to estimate camera parameters and initialize the point clouds for the generated videos of these methods. Thanks to the capability of generating RGB-D content, our results can be directly used in 3DGS reconstruction.

In Table~\ref{tab:scene}, our reconstruction results with VGGT post-hoc outperform the compared baselines, indicating that our generated videos are more consistent in aspect of geometry. The results are even better when initializing point clouds with our own depth output, which demonstrates the effectiveness of our depth generation for scene reconstruction. The qualitative results in Figure~\ref{fig:method} illustrate the same conclusion. Particularly in the last case, our method retains most details of the chandelier, while baseline methods even fail to reconstruct a basic shape.

\subsection{World Generation}\label{subsec:world}
Besides the in-domain comparison on RealEstate, we test Voyager on WorldScore~\cite{duan2025worldscore} static benchmark on world generation. WorldScore consists of 2,000 static test examples that span diverse worlds, \eg, indoor and outdoor, photorealistic and stylized. In each example, an input image and a camera trajectory are provided. The metrics evaluate the controllability and quality of generation, and an average score is presented to show the overall performance. We compare six top methods in the existing benchmark, including two 3D methods WonderJourney~\cite{yu2023wonderjourney} and WonderWorld~\cite{yu2024wonderworld}, and four video methods EasyAnimate~\cite{xu2024easyanimate}, Allegro~\cite{allegro2024}, Gen-3~\cite{gen3}, and CogVideoX~\cite{yang2024cogvideox}.

The scores are reported in Table~\ref{tab:worldscore}. Voyager achieves the highest score on this benchmark. The score shows that our method has a competitive performance on camera control and spatial consistency, compared with 3D-based methods. Our subjective quality is the highest among all methods, further demonstrating the visual quality of our generated videos. Notably, since our video condition is constructed with metric depth, the camera movement in our results are larger than other methods, which is much harder to generate.

\begin{figure*}
    \centering
    \includegraphics[width=0.85\textwidth]
    {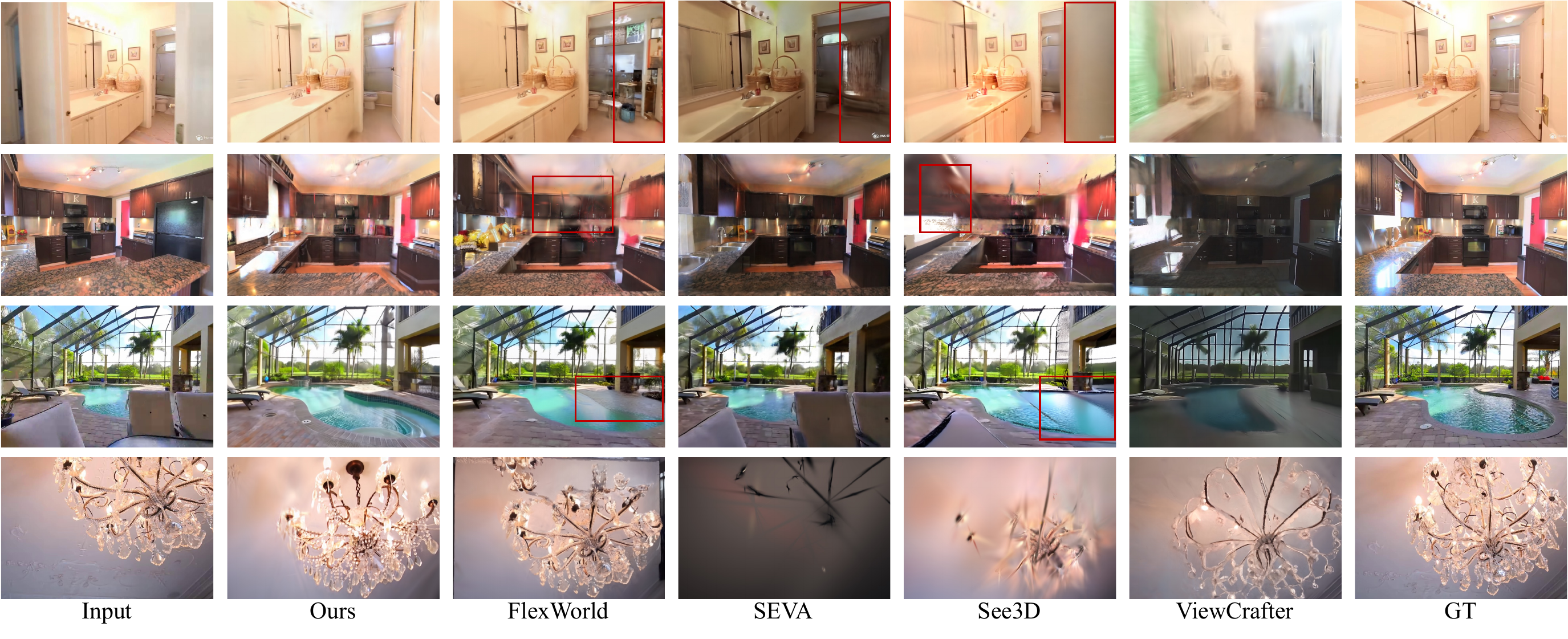}
    \vspace{-1em}
    \caption{Qualitative results on Gaussian Splatting reconstruction. Our results present much more details than the compared baselines. }
    % \vspace{-1em}
    \label{fig:world}
\end{figure*}
\begin{table*}[t]
\centering
\caption{Quantitative comparison on \textit{WorldScore Benchmark}. \textbf{\underline{Bold and underline}} indicates the 1st, \textbf{Bold} indicates the 2nd, \underline{underline} indicates the 3rd.}
\vspace{-0.5em}
\scalebox{0.9}{
\begin{tabular}{l|c|ccc|cccc}
\toprule
\textbf{Method} & \makecell{WorldScore \\ Average} & \makecell{Camera \\ Control} & \makecell{Object \\ Control} & \makecell{Content \\ Alignment} & \makecell{3D \\ Consistency} & \makecell{Photometric \\ Consistency} & \makecell{Style \\ Consistency} & \makecell{Subjective \\ Quality} \\
\Xhline{0.5pt}
WonderJourney &  \underline{63.75}&  \underline{84.6}&  37.1&  35.54&  80.6& 79.03& 62.82 & \textbf{66.56} \\
WonderWorld & \textbf{72.69}&  \textbf{\underline{92.98}}&  51.76&  \textbf{\underline{71.25}}&  \textbf{\underline{86.87}}&  85.56& 70.57& 49.81 \\
\hline
EasyAnimate &  52.85&  26.72&  54.5&  50.76&  67.29&  47.35&  \underline{73.05}& 50.31  \\
Allegro &  55.31&  24.84&  \underline{57.47}&  \underline{51.48}&  70.5&  69.89&  65.6&  47.41  \\
Gen-3 &  60.71&  29.47&  \textbf{62.92}&  50.49&  68.31&  \textbf{87.09}& 62.82 & \underline{63.85}\\
CogVideoX-I2V & 62.15&  38.27&  40.07&  36.73&  \textbf{86.21}&  \textbf{\underline{88.12}}&  \textbf{83.22}& 62.44\\
\hline
\rowcolor{gray!15}
\textbf{Voyager} &  \textbf{\underline{77.62}}&  \textbf{85.95}&  \textbf{\underline{66.92}} &  \textbf{68.92}&  \underline{81.56}&  \underline{85.99}& \textbf{\underline{84.89}}& \textbf{\underline{71.09}}\\
\bottomrule
\end{tabular}
}
\label{tab:worldscore}
\vspace{-0.5em}
\end{table*}

\subsection{Ablation Studies}~\label{subsec:ablation}
To verify the effectiveness of our proposed designs, we conduct ablation studies on our world-consistent video diffusion and long-range world exploration. 

\begin{figure}
    \centering
    \includegraphics[width=\linewidth]{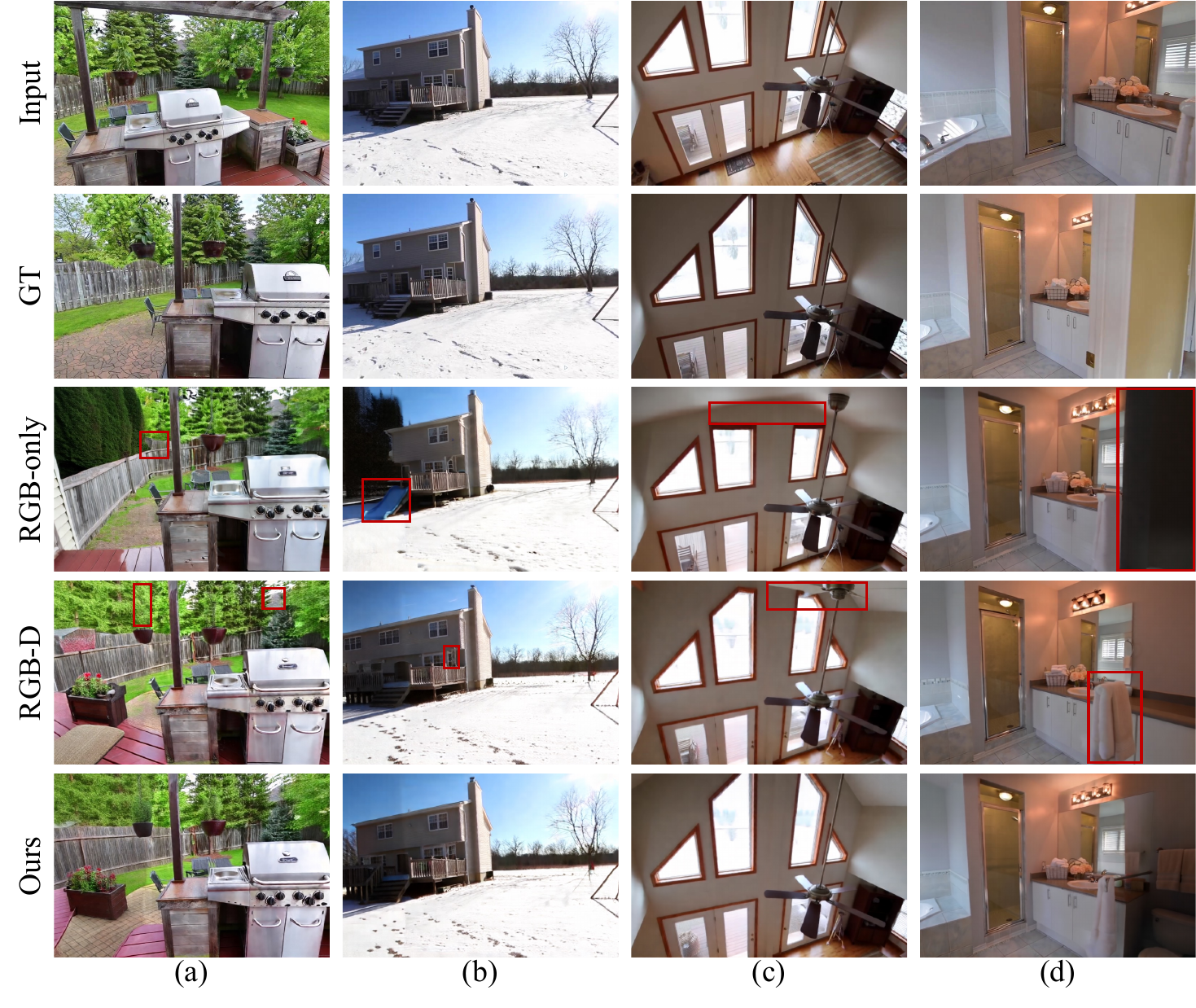}
    \vspace{-1em}
    \caption{Qualitative results on ablation study. We compare the video models in our three training stages. Our final model achieves the highest quality.}
    \label{fig:ablation}
    \vspace{-1em}
\end{figure}

\noindent\textbf{World-Consistent Video Diffusion}
We evaluate our video models trained in the three stages separately on Worldscore benchmark, \ie, (a) model trained only on RGB conditions, (b) model trained on RGB-D conditions, and (c) model attached with additional control blocks. As shown in Table~\ref{tab:ablation}, fusing depth conditions in training can significantly enhance the capability of camera control. The control blocks can further improve the spatial consistency of generated results. We also provide qualitative results in Figure~\ref{fig:ablation}. The RGB-only model may generate inconsistent content when the camera moves to an unseen region. The results of RGB-D model is more consistent with the input image, but it could still produce some minor artifacts. Our final model generates the most reasonable results.

\begin{table}[t]
\centering
\caption{Ablation study on Worldscore.  }
\vspace{-1em}
% \footnotesize
\scalebox{0.9}{
\begin{tabular}{lccccc}
\toprule
\textbf{Metric} & \makecell{Camera \\ Control} & \makecell{Content \\ Alignment} & \makecell{3D \\ Consistency} \\
\midrule
Ours (RGB-only)  & 74.98 & 48.92 & 68.86 \\
Ours (RGB-D) & 85.04 & 65.72 & 78.58 \\
Ours (full) & \textbf{85.95} & \textbf{68.92} & \textbf{81.56} \\
\bottomrule
\end{tabular}
}
\label{tab:ablation}
\vspace{-1em}
\end{table}

\begin{figure}
    \centering
    \includegraphics[width=\linewidth]{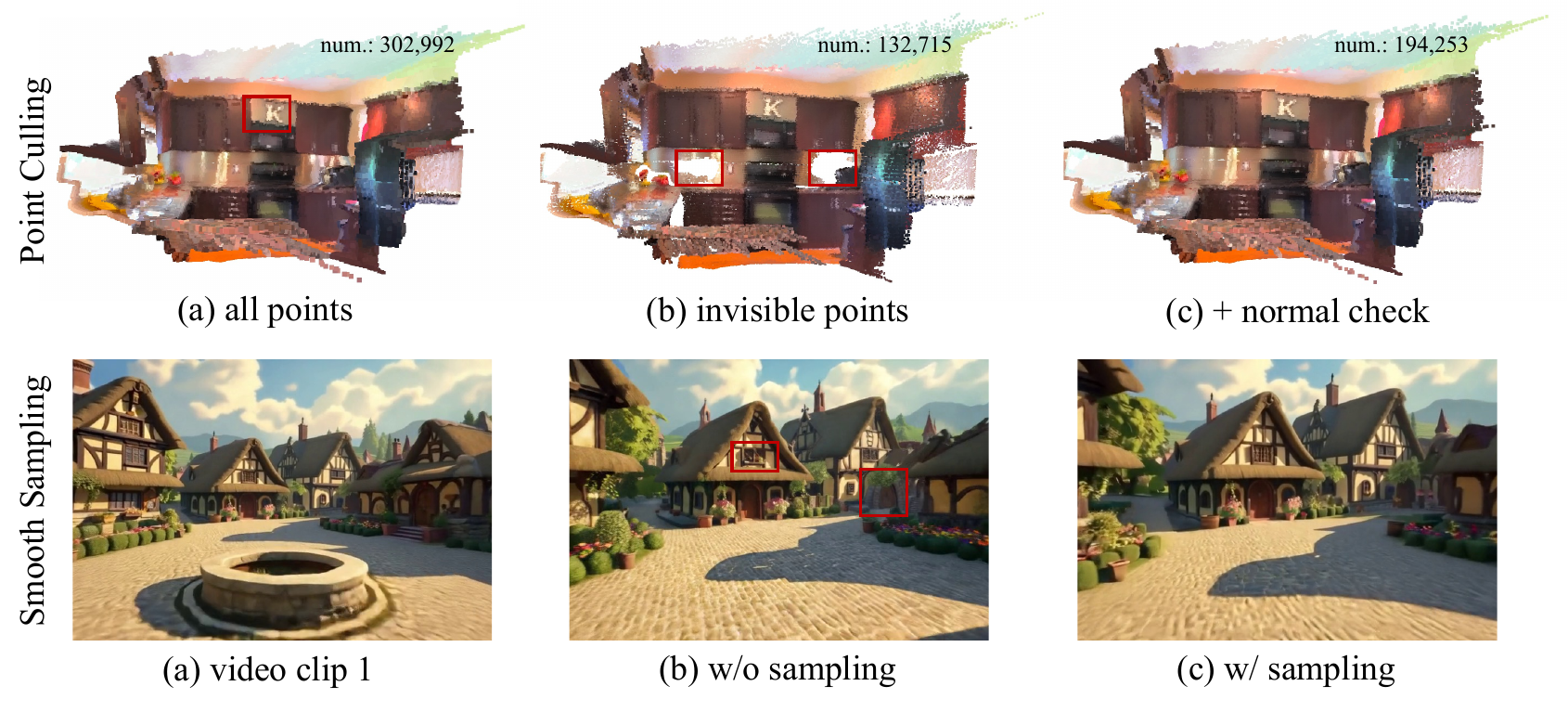}
    \caption{Qualitative results on ablation study. We compare the video models in our three training stages. Our final model achieves the highest quality.}
    \label{fig:ablation_ar}
\end{figure}

\noindent\textbf{Long-range video generation.}
We evaluate the quality of point culling and smooth sampling in Figure~\ref{fig:ablation_ar}. For point culling, storing all points introduces noise, while storing points in the invisible region is insufficient. Results with additional normal check have comparable visual performance with storing all points, but save almost 40\% storage. For smooth sampling, the video clip without sampling may exhibit inconsistencies compared to the first clip. Smooth sampling ensures a seamless transition between two consecutive segments.

\section{Application}\label{subsec:application}
Benefiting from our depth-fused video generation, Voyager supports various 3D-related applications. 

\begin{figure*}[h]
    \centering
    \includegraphics[width=0.9\textwidth]{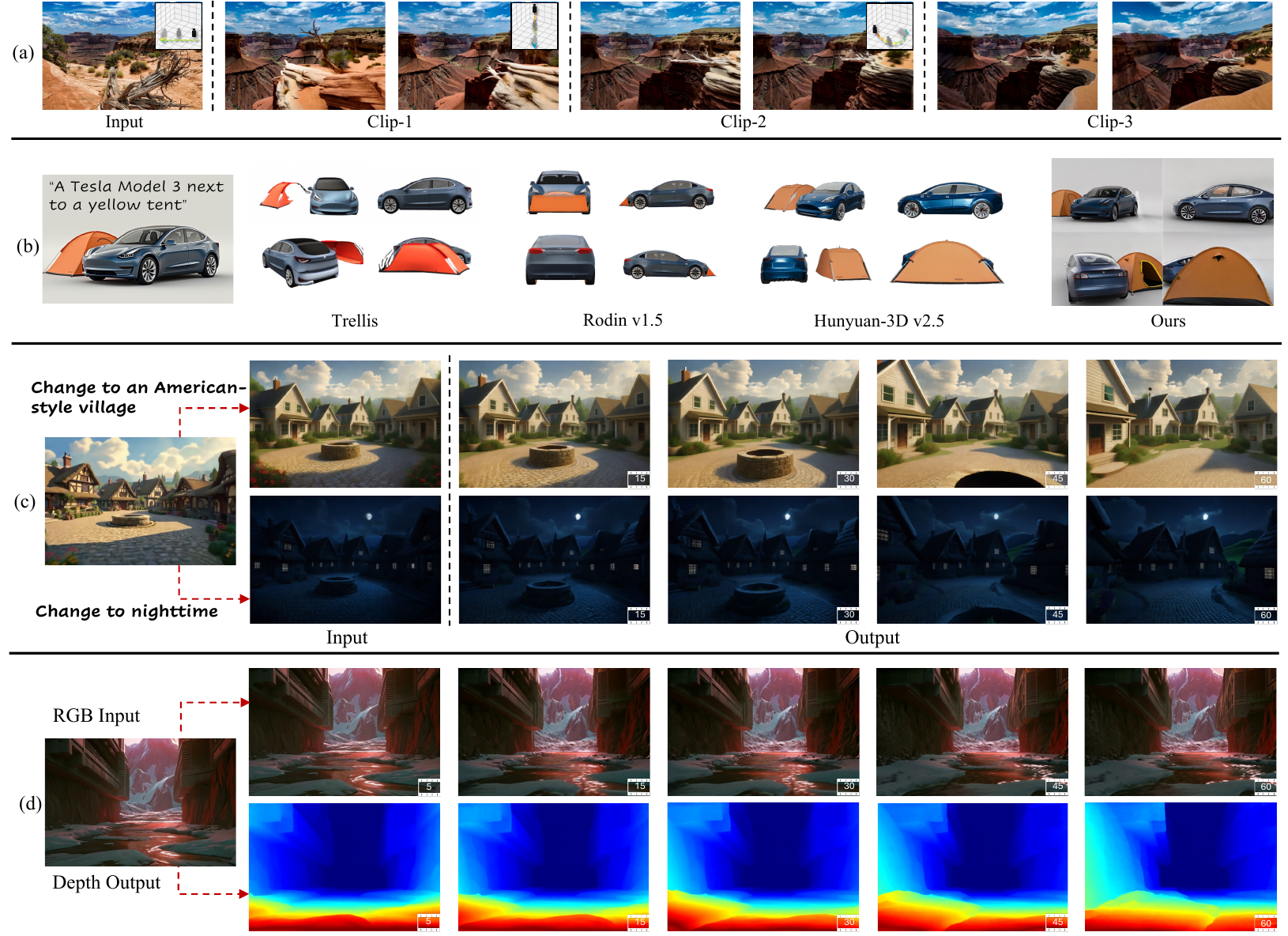}
    \caption{Applications: (a) Long-range video generation. (b) Image-to-3D generation. (c) World-consistent video style transfer. (d) Monocular video depth estimation.}
    \label{fig:application}
\end{figure*}

\noindent\textbf{Long Video Generation.} As explained in Sec.~\ref{subsec:infer}, our method allows long-range video generation with efficient world caching and smooth video sampling. In Figure~\ref{fig:application}(a), we provide an example consisting of three video clips, with totally different camera trajectories among clips. The results present camera controllability and spatial consistency of the generated video, demonstrating that our method is capable of long-range world exploration.

\noindent\textbf{Image-to-3D Generation.} Native 3D generative models can hardly handle the generation of multiple objects. In Figure~\ref{fig:application}(b), we use three state-of-the-art 3D generation methods Trellis~\cite{xiang2024structured}, Rodin v1.5~\cite{rodin}, and Hunyuan-3D v2.5~\cite{hunyuan} to generate a simple combination where a car leans against a tent. Rodin failed to generate the tent, while Trellis produced a tent with missing parts. Hunyuan successfully generated two complete objects, but the spatial relationship was inaccurate, with the tent being too far from the car. Our method not only generates the correct content, but also produces more realistic visual effects. The tent is even visible through the car window in the side view.

\noindent\textbf{Depth-Consistent Video Transfer.} Generating a spatially consistent video with a different style typically requires training a stylized video model. However, to achieve the desired effect with our model, we only need to replace the reference image while retaining the original depth condition. As shown in Figure~\ref{fig:application}(c), we can change the original video to American-style or to the night.

\noindent\textbf{Video Depth Estimation.} Our video model is naturally capable of estimating video depth. In Figure~\ref{fig:application}(d), our predicted depth can preserve the details on the architectures.

\section{Conclusion}
In this paper, we present \textbf{Voyager}, a world-consistent video generation framework for long-range world exploration. The proposed RGB-D video diffusion model can produce spatially consistent video sequences that align with the input camera trajectories, allowing direct 3D scene reconstruction. This supports auto-regressive and consistent world expansion.  Experiments demonstrate high visual fidelity and strong spatial coherence in both generated videos and point clouds.


\bibliographystyle{ACM-Reference-Format}
\bibliography{sample-base}

\clearpage
\appendix
In this supplement, we will introduce more details of training implementation (Sec.~\ref{suppl:impl}), our video diffusion model (Sec.~\ref{suppl:method}), and our video data engine (Sec.~\ref{suppl:data}). Finally, we provide more generation results in Sec.~\ref{suppl:more}.

\section{Implementation Details}\label{suppl:impl}
Our training basically follows the image-to-video model of HunYuan-Video~\cite{kong2024hunyuanvideo}. We divide the training into three stages: the first stage only trains the RGB video model; in the second stage, depth is introduced into the training; and in the third stage, the DiT parameters are frozen and ControlNet blocks are incorporated for training. We use all three datasets in the first training stage. However, DL3DV is removed in the second stage due to its fast camera motion, which makes it unsuitable for depth training. In the third stage, we train solely on the UE dataset with its ground-truth depth. During training, we randomly select a width-height ratio from $[1, 1.25, 1.5, 1.75]$ to support the generation of videos with multiple aspect ratios. The number of generation frames for a single pass is 49.

\begin{figure}[h]
    \centering
    \includegraphics[width=0.8\linewidth]{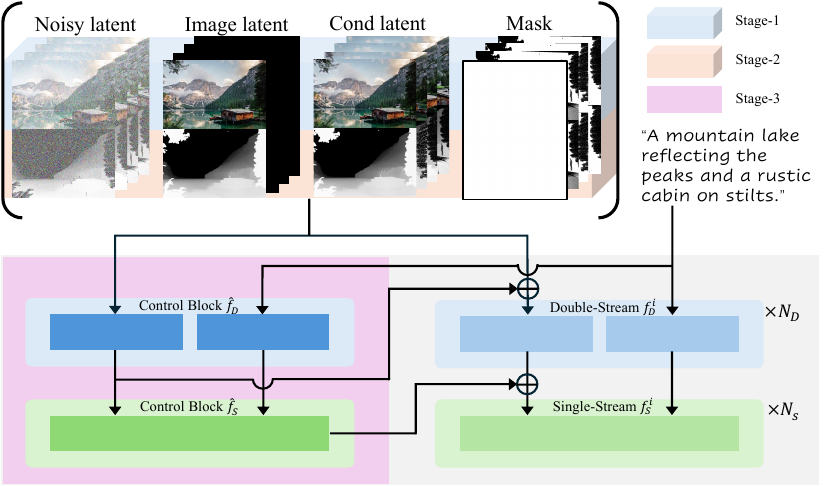}
    \caption{Details of world-consistent diffusion model.}
    \label{fig:dit}
    \vspace{-2em}
\end{figure}

\section{World-Consistent Video Diffusion}\label{suppl:method}
We provide the details of our video diffusion model in Figure~\ref{fig:dit}. The input of the diffusion model includes noisy latents $\mathbf{z}_t$, input image latents $\mathbf{z}_0^r$, condition latents $\hat{\mathbf{z}}_0$, and down-sampled mask $m$. To align the temporal dimension, we pad $\mathbf{z}_0^r$ with zero latents. In the first stage of training, only the RGB-related latents are concatenated in the channel dimension and are then fed to the diffusion model. In the second stage, we inject depth-related latents into the input. We fine-tune the parameters of the original diffusion structure in the first two stages, two additional Transformer blocks are trained in the final stage. The aggregated features in these two blocks are added back on a pixel-wise basis.

\section{Scalable Video Data Engine}\label{suppl:data}
\begin{figure}
    \centering
    \includegraphics[width=\linewidth]{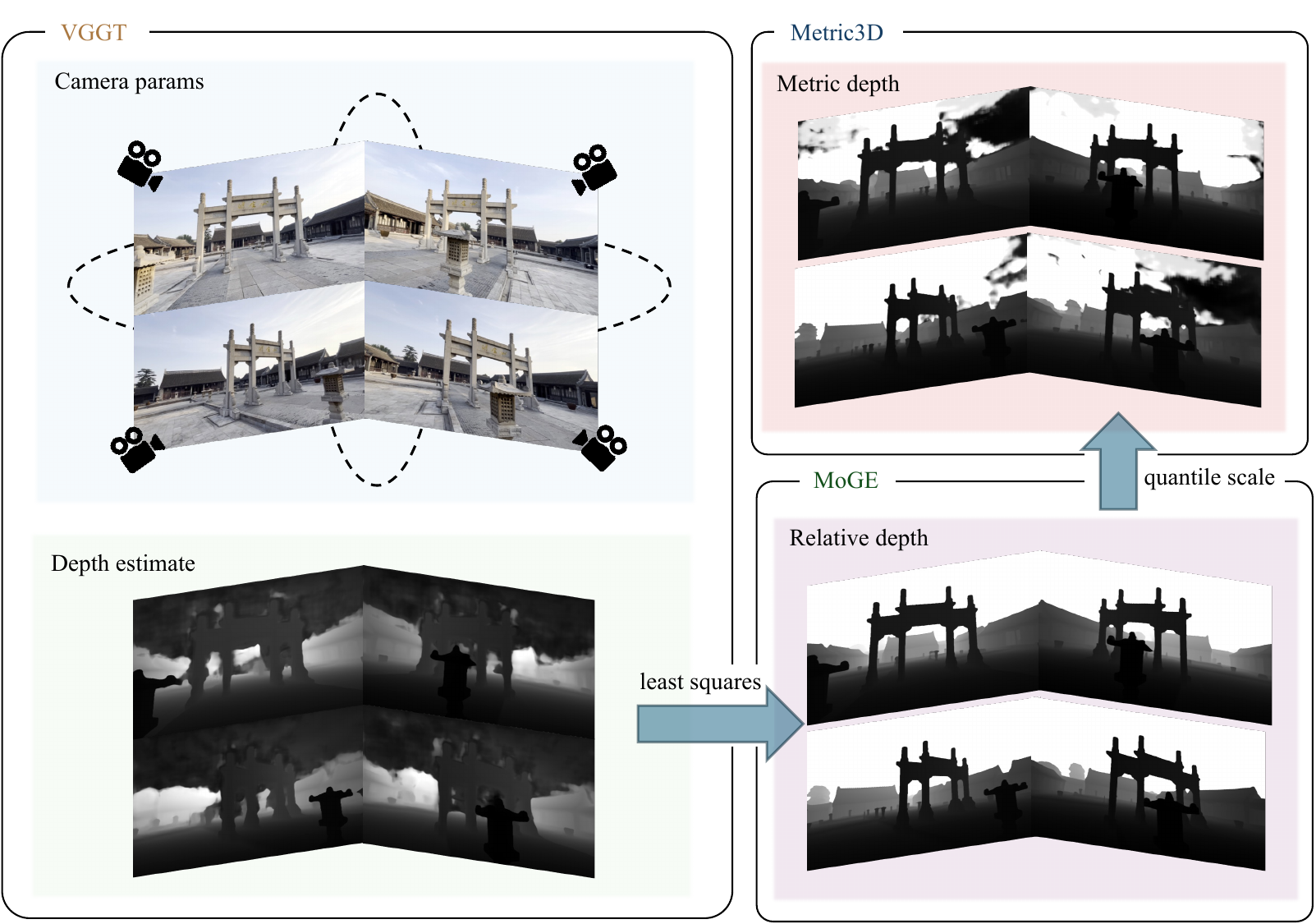}
    \caption{Overview of our scalable video data engine.}
    \label{fig:data_engine}
\end{figure}

\begin{figure}
    \centering
    \includegraphics[width=\linewidth]{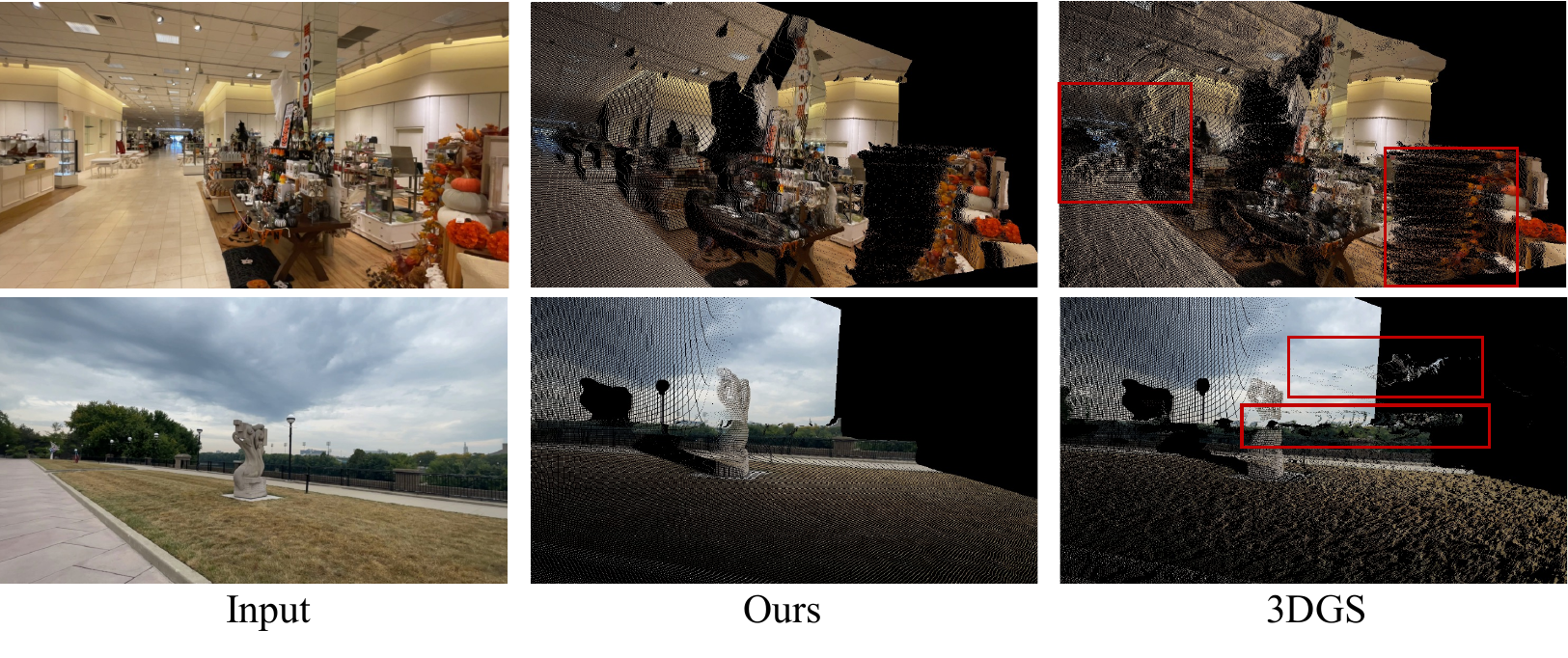}
    \caption{Training warp images compare.}
    \label{fig:moge_3dgs_warp}
\end{figure}

Accurate camera parameters and depth are crucial for model training. As shown in Fig. \ref{fig:data_engine}, we propose a more robust data processing pipeline. Compared to Flexworld~\cite{chen2025flexworld}, since our depth estimation method is more consistent and accurate than the depth rendered by 3DGS, our warped images are more precise, as shown in Fig. \ref{fig:moge_3dgs_warp}.

Specifically, we first use VGGT~\cite{wang2025vggt} to estimate camera parameters and depth for all video frames. The depth estimated by VGGT is not accurate enough, but it is aligned with camera poses. 
To further improve the depth estimation, we then employ MoGE~\cite{wang2024moge} as a robust depth estimator. Specifically, we first convert the depth into disparity, and then use a least squares-based optimization strategy to minimize the disparity difference between the depth frames generated by
VGGT and those generated by MoGE.
The optimization is represented as:
\begin{equation}
    \min_{scale, bias} \left\| \mathbf{M} \cdot \left( \frac{scale}{d_{MoGE}} + bias - \frac{1}{d_{VGGT}} \right) \right\|^2,
\end{equation}
where \( scale \) and \( bias\) represent the scale and shift factors respectively. The mask \( \mathbf{M} \) represents the valid non-sky regions.

Finally, to ensure scale uniformity across datasets, we estimate the metric depth range using Metric3D~\cite{hu2024metric3d} and map the estimated depths into this range. 
\begin{equation}
s_{\text{metric}} = \frac{q(0.8,\mathbf{d}_{\text{Metric3D}}) - q(0.2,\mathbf{d}_{\text{Metric3D}})}{q(0.8,\mathbf{d}_{\text{MoGE}}) - q(0.2,\mathbf{d}_{\text{MoGE}})} 
\end{equation}
\begin{equation}
    d_{\text{metric}} = s_{\text{metric}} \cdot d_{MoGE}
\end{equation}
\begin{equation}
    C_{cam}^{metric} = \left[
    \begin{array}{cc}
    R & s_{\text{metric}} \cdot T \\
    0 & 1
    \end{array}
    \right]
\end{equation}
where \( q(p, \mathbf{x}) \) represents the \(p\)-th quantile of vector \( \mathbf{x} \), while \( d_{metric}\) and \(C_{cam}^{metric} \) denote the final metric depth and camera extrinsics, respectively.

\section{More Results}\label{suppl:more}
\begin{figure}
    \centering
    \includegraphics[width=\linewidth]{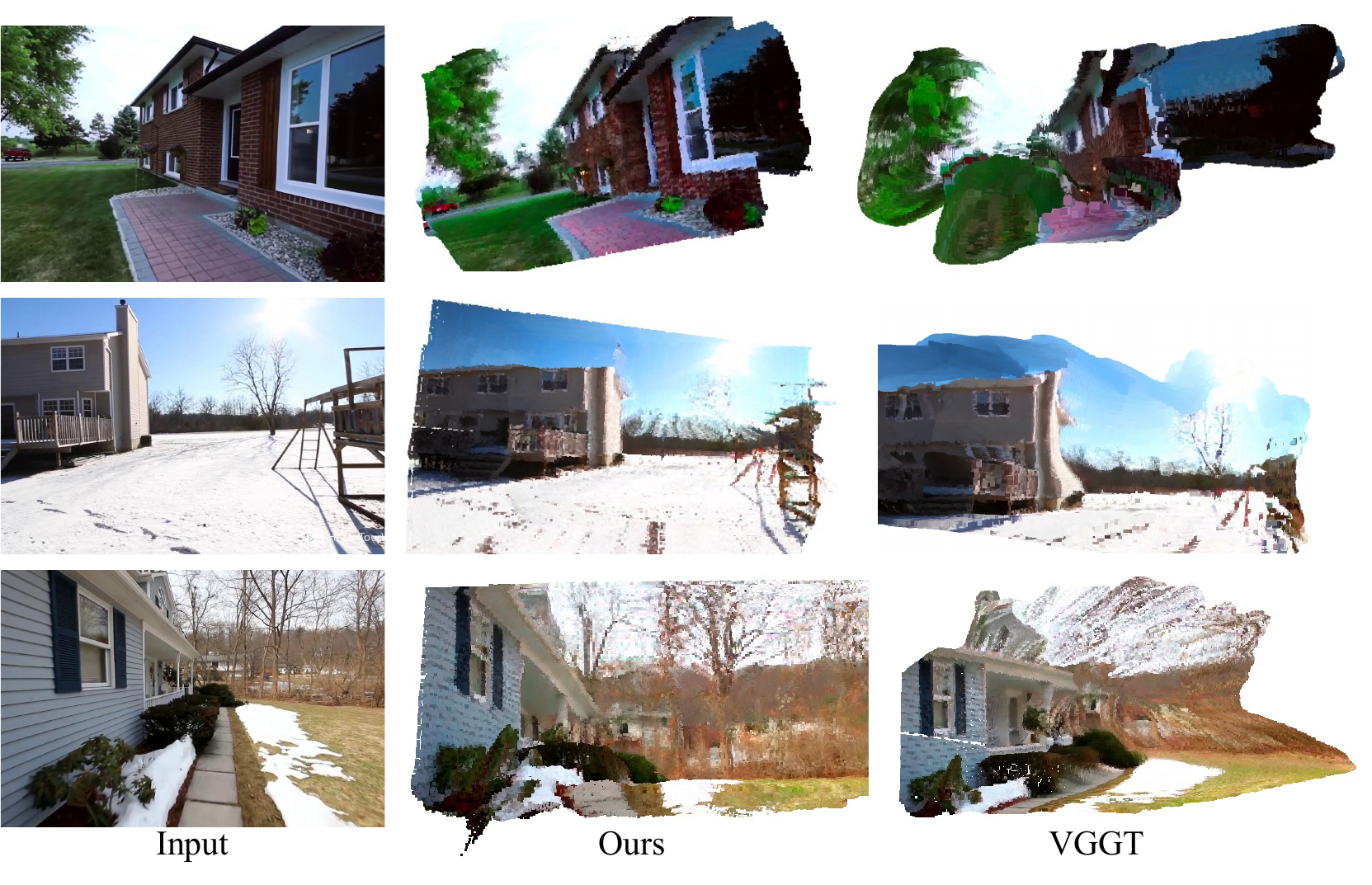}
    \caption{Comparison of initialization point clouds of ours and VGGT.}
    \label{fig:pcd_compare}
\end{figure}

We provide visualization results for the initialization of 3D reconstruction in Figure~\ref{fig:pcd_compare}. Our point cloud results are much better than VGGT, demonstrating that our depth estimation is more accurate than VGGT.
 
\begin{figure*}
    \centering
    \includegraphics[width=\textwidth]
    {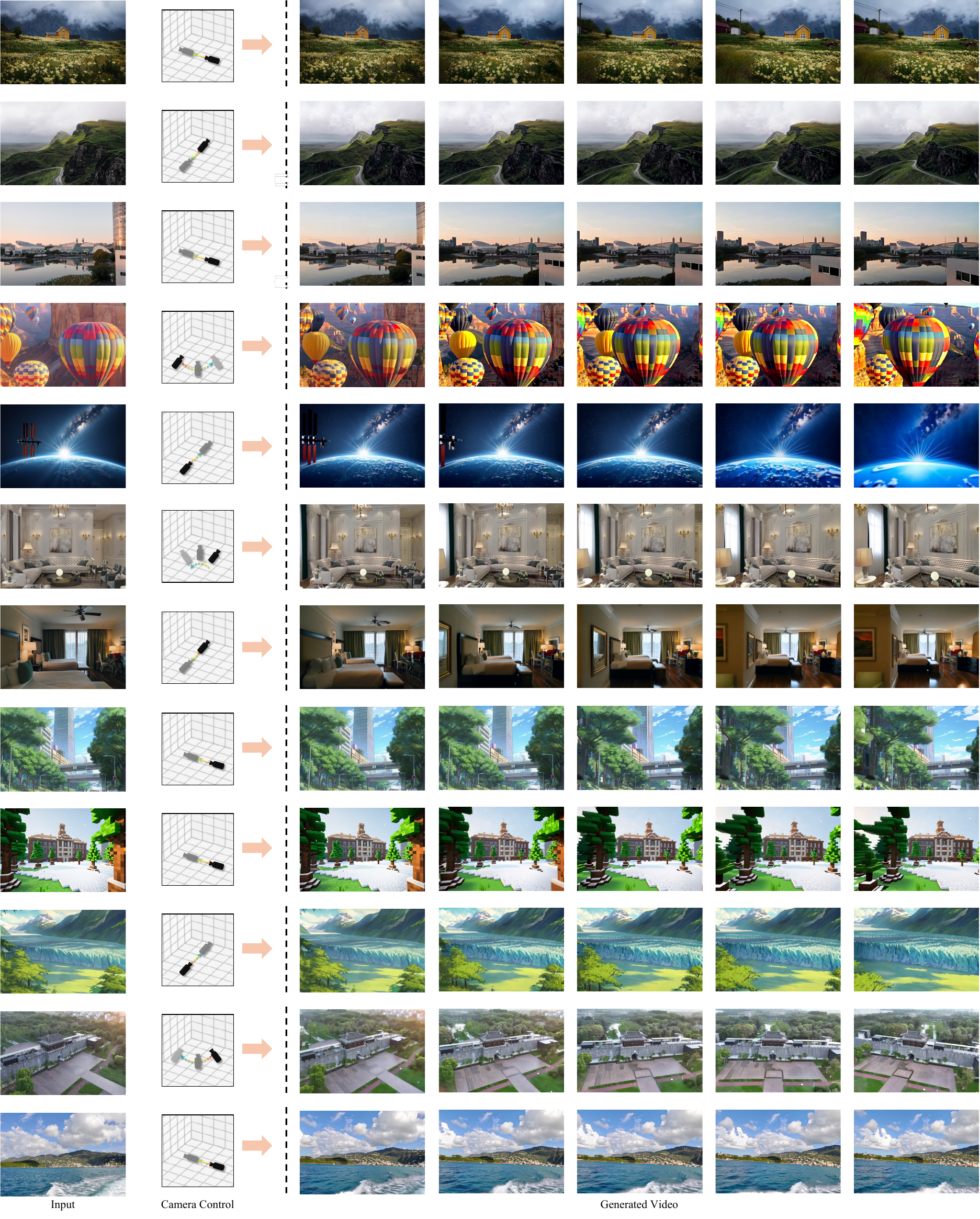}
    \caption{More Results.}
    \label{fig:more_results1}
\end{figure*}

\begin{figure*}
    \centering
    \includegraphics[width=\textwidth]
    {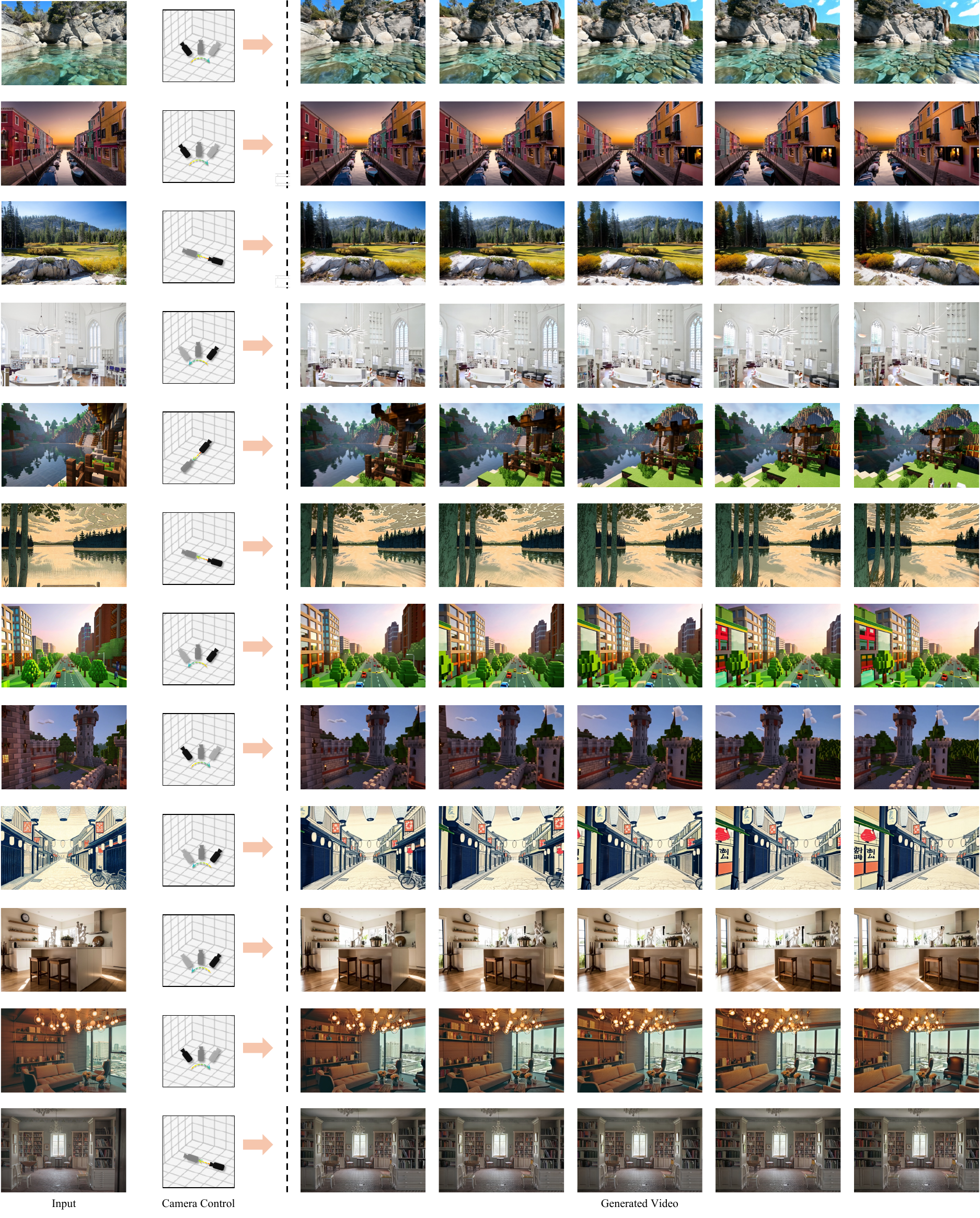}
    \caption{More Visualization Results.}
    \label{fig:more_results2}
\end{figure*}

We also provide more generation results in Figure~\ref{fig:more_results1} and Figure~\ref{fig:more_results2}.


\end{document}